# A comparison between Recurrent Neural Networks and classical machine learning approaches In Laser induced breakdown spectroscopy

Fatemeh Rezaei [1*], Pouriya Khaliliyan [1,2], Mohsen Rezaei [3], Parvin Karimi [4], Behnam Ashrafkhani [5]

[1] Department of Physics, K. N. Toosi University of Technology, Tehran, Iran

[2] PDAT Laboratory, Department of Physics, K. N. Toosi University of Technology, Tehran, Iran

[3] Department of Industrial Engineering, University of Science and Technology of Mazandaran, Behshahr, Iran

[4] Department of Physics, South Tehran Branch, Islamic Azad University, Tehran, Iran

[5] Department of Physics, University of Calgary, Canada, Calgary, Alberta

*Corresponding author*: fatemehrezaei@kntu.ac.ir

## Abstract

Recurrent Neural Networks are classes of Artificial Neural Networks that establish connections between different nodes form a directed or undirected graph for temporal dynamical analysis. In this research, the laser induced breakdown spectroscopy (LIBS) technique is used for quantitative analysis of aluminum alloys by different Recurrent Neural Network (RNN) architecture. The fundamental harmonic (1064 nm) of a nanosecond Nd:YAG laser pulse is employed to generate the LIBS plasma for the prediction of constituent concentrations of the aluminum standard samples. Here, Recurrent Neural Networks based on different networks, such as Long Short Term Memory (LSTM), Gated Recurrent Unit (GRU), Simple Recurrent Neural Network (Simple RNN), and as well as Recurrent Convolutional Networks comprising of Conv-SimpleRNN, Conv-LSTM and Conv-GRU are utilized for concentration prediction. Then a comparison is performed among prediction by classical machine learning methods of support vector regressor (SVR), the Multi Layer Perceptron (MLP), Decision Tree algorithm, Gradient Boosting Regression (GBR), Random Forest Regression (RFR), Linear Regression, and k-Nearest Neighbor (KNN) algorithm. Results showed that the machine learning tools based on Convolutional Recurrent Networks had the best efficiencies in prediction of the most of the elements among other multivariate methods.

*Keywords*: LIBS, Recurrent Neural Networks, Concentration prediction, Machine learning, LSTM, GRU, SimpleRNN, Convolutional Recurrent Networks.

## Introduction

Aluminum is one of the most abundant metals on the earth, and the most widespread element in the earth's crust, after silicon and oxygen. The main properties of aluminum are its low density, high heat conductivity, ductility, corrosion resistance, being a catalyst, low temperature resistance, high reflectivity, and sound absorbing. Studying the characteristics of aluminum's concentration can be performed by different methods. Laser induced breakdown spectroscopy (LIBS) as a type of atomic emission spectroscopy is a fast, online optical technique that employs a highly energetic laser pulse as the excitation source. These advantages make LIBS a powerful tool for the quantitative analysis of different materials, especially metal samples [1,2]. In this method, the laser is focused to produce hot plasma, which atomizes and excites the targets. Notice that the formation of plasma only starts when the focused laser energy reaches a certain threshold for optical breakdown. Then the emissions of plasma are conducted to a spectrometer for identification of the constituent elements of the analyzed sample [3-4].

Various research groups have studied the classical [5-14] and deep machine learning algorithms [15-18] in LIBS technique for improving the state of the art quantitative estimations by using simple Artificial Neural Networks (ANN) [19-23], Convolutional Neural Network [15,18,19], and so on. For instance, Xu et al. [15] used the algorithm of Convolutional Neural Network for studying the LIBS spectra of two-dimensional soil samples. They demonstrated that Convolutional Neural Networks outperformed traditional preprocessing approaches by preventing overfitting. They presented that the multi-tasking of Convolutional Neural Network enhanced the prediction because of its capacity to learn inherent structures. Zhao et al. [16] employed the Convolutional Neural Network methodology for the prediction of 16 types of brand iron ores from Australia, Brazil, and South Africa. They showed that the Convolutional Neural Network outperformed common machine learning methods. In addition, they explained that the effectiveness of Convolutional Neural Network assisted LIBS was interpretation by t-SNE and chemical components of iron ores. Chen et al. [17] used label-free laser-induced breakdown spectroscopy in order to recognize the tumor cells. They collected the LIBS spectra from cells located on a silicon substrate and employed deep learning algorithms for simultaneous classification of different cancers. Furthermore, they utilized gradient-weighted class activation mapping for the interpretation of the results of the one-dimensional convolution neural network (1D-CNN). They represented that 1D-CNN algorithms attained a mean sensitivity of 94.00%, a mean accuracy of 97.56%, and a mean specificity of 98.47%. Moreover, Ewusi-Annan et al. [24] applied two machine learning techniques, namely PLS and feed-forward ANNs, to earth and Mars data. They have shown that they can forecast the preprocessed spectra of samples with very high accuracy.

This study represents the influences of different Recurrent Networks such as Long Short Term Memory (LSTM), Gated Recurrent Unit (GRU), Simple Recurrent Neural Network (SimpleRNN), and Convolutional Recurrent Networks on LIBS emission spectra analysis. In the current work, the focus is improvement of the precision of the quantitative analysis in LIBS by introducing the best multivariate methodology. Here, a comparison is made with the Support Vector Regressor (SVR), the Multi Layer Perceptron (MLP), Decision Tree algorithm, Gradient

Boosting Regression (GBR), Random Forest Regression (RFR), and the k-Nearest Neighbor (KNN) algorithm for quantitative evaluations in terms of both loss and precision in the LIBS technique. To the best of the authors' knowledge, Convolutional Recurrent Networks were used for the first time in the LIBS technique for prediction, yielding impressive results. Furthermore, a comprehensive comparison among different methods is performed for quantitative analysis. All of these methods are used to quantify the corresponding elements of Si, Fe, Cu, Zn, Mn and Mg in different aluminum standard samples. Here, the mean squared error (MSE), the mean absolute error (MAE), and the mean absolute percentage error (MAPE) are calculated to evaluate the prediction ability of the mentioned statistical models.

**Experimental setup**

A typical experimental setup of the LIBS method shown in Fig. 1 has been utilized in the current work [25,26]. The laser beam for producing plasma is a Q-switched Nd:YAG laser (Continuum, Surelite III, China) at 1064 nm wavelength, with a pulse width of 8 ns, laser energy of 50 mJ, and a repetition rate of 10 Hz. The samples are different aluminum standards supplied from the Razi metallurgical research center in Iran. At a specific delay time, the laser pulse is guided to a beam splitter and divided into two parts. One section is conducted to a photodiode for launching the delay generator. Then the ICCD camera (Andor, iStar DH734, Ireland) receives a pulse from the delay generator to start the acquisition of data. The other part is passed through a λ/2 plate and a Glan–Taylor prism for adjustment of the laser energy. Then, a laser pulse is focused on the surface sample by a 20 cm focal length lens. It should be mentioned that the position of the strike of the laser pulse is varied by a XYZ stage during the experiment.

Spatially integrated plasm a radiations are collected using a quartz lens with focal length of 5 cm, and with the aid of an objective lens, then transmitted to an optical fiber and guided to an Echelle spectrograph (Kestrel, SE200) to obtain spectrally resolved light spectrum. The recorded spectra are temporally analyzed by adjusting of the gate and delay time of the ICCD camera. For the spectral analysis, the acquisition delay time between the laser pulse and plasma radiations is altered. Then, the optimum delay time is chosen for the experimental analysis by testing various delay times and selecting the best one containing the highest signal to noise ratio. In this research, the optimum delay time between the laser pulse and plasma radiation acquisition for the neutral elements of Mn, Cu, Mg, and Si is about 6 μs and for the two ionic elements of Fe and Zn is 2 μs, and the gate width of 5 μs is selected in the whole of the experiments.

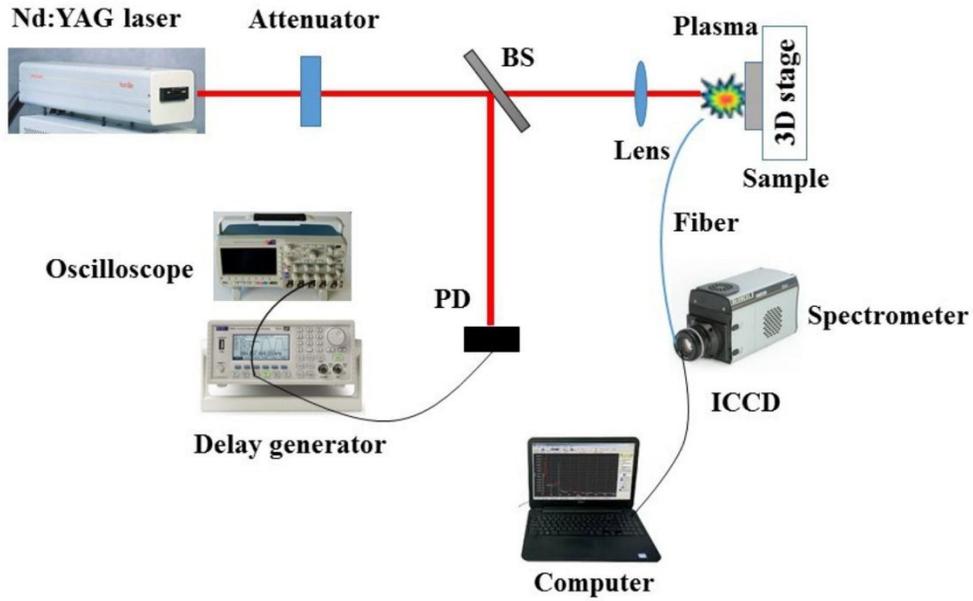

Figure 1. A schematic of the LIBS experimental set-up for obtaining the spectral line's intensities as input of the multivariate algorithms.

## Materials and Methods

### 1. Simple Recurrent Neural Network (SimpleRNN)

Generally, SimpleRNNs are networks which work based on a recurrent architecture. In essence, SimpleRNN return memory to the data, and affect the memory of previous states as the new input. It should be noted that SimpleRNN essentially evoke the dynamic behavior of the networks, but by applying memory, the dynamic behavior of a system which has been operating in the past will be reconstructed. The more complexity in the Neural Network system will results in the more complexity in the dynamics behavior and finally, the more recurrent process in the Recurrent Network, leads to more complex nonlinear equations defined in proportion to the complexity of the model. It can be mentioned that the different steps of RNN are as follows [27]:

1- $x_t$ acts as an input of model at time *t*.

2- $s_t$ or hidden state at time *t* is defined as the memory of the recurrent model. The nonlinear functions of Relu or tanh related to *f* are usually defined as [27]:

$$s_t = f(U\, x_t + W\, s_{t-1}), \tag{1}$$

where, *U*, *W* are weights. $x_t$, is input and $s_t$ is the hidden state to build up the memory, respectively.

3- $o_t$ as the output of the network at time *t* is considered. The probability $softmax$ function for the nonlinear network activation function is used as [27]:

$$o_t = softmax(Vs_t), \tag{2}$$

here, $V$ is the weight.

In this research, different spectral line's intensities are employed as input data with different features of the main spectra, and type of materials for concentration prediction.

## 2. Long Short Term Memory (LSTM)

LSTM is an extended form of the RNN Networks. LSTM uses memory blocks instead of RNN building units to solve the problem of gradient vanishing. This network can memorize long-term dependencies better than SimpleRNN, which means that useful information is kept in memory and construct memory. LSTM works in such a way that it can delete unnecessary information of the memory and retain useful information. This algorithm contains different gates as below:

1- $f_t$ in the forget gate decides whether or not store the data in memory depending on its importance as [28]:

$$f_t = \sigma(w_i.[h_{t-1}, x_t] + b_i), \tag{3}$$

here, $\sigma$ is the activation function, $b_i$ is bias and $x_t$, $h_{t-1}$ are the input vector, and output result to the memory cell at the considered time, respectively.

2- Input gate for storing data must be defined as [28]:

$$i_t = \sigma(w_i.[h_{t-1}, x_t] + b_i). \tag{4}$$

$$\tilde{c}_t = \tanh(w_c[h_{t-1}, x_t] + b_c), \tag{5}$$

here, $i_t$, $\tilde{c}_t$, $w_c$ are the cell activation state and its weights, and $b_c$ is the bias magnitude.

3- New data with $c_t$ must be added to the stored data in memory at this point. Then, the network decides whether to save or delete the data as below [28]:

$$c_t = f_t * c_{t-1} + i_t * \tilde{c}_t, \tag{6}$$

4- Finally, the network must decide what to print as output [28]:

$$o_t = \sigma(w_o[h_{t-1}, x_t] + b_o). \tag{7}$$

$$h_t = o_t * \tanh(c_t), \tag{8}$$

here, $o_t$ is the output gate.

## 3. Gated Recurrent Unit (GRU)

Gated recurrent unit or GRU is an algorithm with a Recurrent architecture similar to LSTM, which is designed to solve the gradient vanishing problem. It should be mentioned that GRU has fewer parameters than LSTM, and also has no output port. Generally, the GRU has two gates based on what it decides to store and update, and as well as comprises below gates as follows:

i) At the reset gate, the network decides how much information must to forget [29]:

$$r_t = \sigma(w_r * [h_{t-1}, x_t] + b_r), \tag{9}$$

ii) At the update gate, the network decides what information must be deleted and what information must be transferred as the output data and finally must be updated as [29]:

$$z_t = \sigma(w_z * [h_{t-1}, x_t] + b_z). \tag{10}$$

$$\check{h}_t = \tanh(w_h * [h_{t-1}, x_t] + b_h). \tag{11}$$

$$h_t = (1 - z_t) * h_t + z_t * \tilde{h}_t, \tag{12}$$

In above equation, $z_t$ is the update gate and $r_t$ is the reset gate, respectively.

## 4. Convolution Neural Network (CNN)

Convolutional Neural Networks (CNNs) are a special type of Multi-Layer Perceptron (MLP) algorithm. CNNs are used to filter data and minimize the data preprocessing. They are basically composed of three layers, the first layer is the convolution layer, the second layer is the pooling and the third layer is a fully connected layer which has the *ReLU* activation function. The convolution layer applies the convolution operation on the input data with the $w \in R^{fd}$ filter and creates a feature map, where $f$ represents the feature of the input data so that the output is generated, and the data set is transferred into the input.

A new feature mapping $f_m$ is obtained from the features set of $f$ as follows [30]:

$$hl_i^{fm} = \tanh(w^{fm} x_{i:i+f-1} + b). \tag{13}$$

here, x is network input vector. It should be mentioned that the filter $hl$ is defined for each set of features $f$ in the input data, so that a feature map is formed as hl, where $b$ is the bias quantity.

Then, the output of the convolution layer is given to the pooling layer and the convolution layer uses the *ReLU* activation function. In the pooling layer, the down-sampling operation is performed and the max-pooling operation is used for each feature mapping. Moreover, the extracted features are utilized as the input of the fully connected layer. It must be considered to determine the final output of the network. It should be noted that the details of these algorithms are explained in Ref. [31].

## 5. Support Vector Regressor (SVR)

Support vector regressor (SVR) method is a kernel-based regression algorithm which is constructed according to the principle of support vector machine (SVM). SVR methodology is characterized as a powerful technique for function estimation and pattern recognition. A training dataset of *n* points are given in the form $(x_1,y_1), \ldots (x_n,y_n)$, where, $x_n$ indicates the input spectrum, $y_i$ denotes the intensities related to the target value, and n corresponds to the number of samples.

Generally, hyperplane in SVM method is a separating hyperplane between two data sets, but in SVR model this is the line which is exploited to predict the continuous output. Here, margin is a region bounded between two hyperplanes. The main idea is minimizing error, individualizing the hyperplane that maximizes the margin, by considering that part of the error tolerated [7].

It should be mentioned that, SVR model maps the dataset from the nonlinear low dimensional space to linear high-dimensional with the kernel function usage so that the nonlinear data convert to the linear data in a new coordinate. In summary, SVR can change the nonlinear relationship of input datasets with the aid of different kernel functions [32].

The main goal of SVR is finding a function f(x) with a deviation from $y_n$, for each training point *x*, by a value not greater than $\varepsilon$. Each hyperplane can be written as the set of points *x* obeying as flat as possible for f(x) function [33]:

$$f(x) = wx + b. \tag{14}$$

where, *w* is the (not necessarily normalized) normal vector to the hyperplane and *b* is the bias term.

SVR regression model is formulated as minimization of the below functional equation [10, 34]:

$$\frac{1}{2}\|w\|^2 + C\sum_{i=1}^{l}(\xi_n + \xi_n^*). \tag{15}$$

subject to 29:

$$\begin{bmatrix} y_n - (wx_n + b) & \leq & \varepsilon + \xi_i \\ (wx_n + b) - y_n & \leq & \varepsilon + \xi_i^* \\ \xi_n, \xi_n^* & \geq & 0 \end{bmatrix} \tag{16}$$

here, $\xi_n$ and $\xi_n^*$ are two positive slack variables for measuring the deviation. *C* is a box constraint constant with a positive value which controls a lot of imposed on the observations located outside of epsilon margin ($\varepsilon$) and help to avoid overfitting.

In other word, linear SVR in dual formula obeys a Lagrangian function constructed from primal function by considering nonnegative multipliers $\alpha_n$ and $\alpha_n^*$ for every observation $x_n$ as:

$$y = \sum_{n=1}^{l}(\alpha_n - \alpha_n^*) \cdot (x_n x) + b. \tag{17}$$

## 6. Multi Layer Perceptron (MLP)

Multilayer Perceptron (MLP) is neural network architecture, which have been used to build the predictive regression model. In the training area of this algorithm, the back-propagation algorithm is employed to calculate the gradient. On the other hand, to perform learning with utilizing this gradient, another algorithm is used as an optimization function. In this research, SGD optimization function is applied as an optimization function.

MLP is a Feed Forward network containing several hidden layers between the input and output layers. First, the input x is given to the input layer. Then, the hidden values of z or pre-activation are calculated. Each hidden layer applies a non-linear activation function for weighted summation and then, calculates the loss. Generally, it determines the weights and bias of the network using loss [35].

## 7. Decision Tree Regression

The decision tree algorithm is designed using information theory. It consists of a set of questions and answers following a unique path. Decision tree algorithm starts from a root and divides into nodes, representing different attributes and edges, representing the possible outcomes. Like leaves on the tree, these nodes help to classify the data or make regression in a continuous mode. Therefore, to predict the result, one has to follow this nested hierarchy of attributes (nodes). It should be stressed that on each node, answering the question leads to the next step on the tree. Following the flowchart, one can predict the outcome at the end of the path [36,37].

## 8. k-nearest neighbor's algorithm

The k-nearest neighbor (KNN) algorithm is a powerful nonparametric predicting tool which classifies data according to a similar pattern to assign into the same class. Although it has some drawbacks, such as being noise-sensitive for incomplete data and less efficient working with big data, it is employed in different learning algorithms. In the KNN method, each sample uses its neighbor distance or nearest objects. The sample data is classified into a class in common between nearest neighbors. Hence, the average weight allocated to the predicted value is inversely proportional to the distance. It should be mentioned that the supervised learning occurs for the labeled (training) data and unlabeled (test) data. The task of the learning process can be either for regression or classification with majority voting [38,39].

## 9. Gradient boosting regression (GBR)

The gradient boosting regression or GBR is a practical and robust model in machine learning widely used in different studies. The boosting machine algorithm indeed employs the sequence of the weak learner to minimize the error or loss of the model. First, a weak learner which acts better than a random classifier is inserted into the model. The classifiers have their value which will change after training. The gradient boosting method reduces the weight of the loss function by supervising the weak learners. Then the ensemble on these trained learners generates a nonlinear function which classifies the problems and solves the regressions [35, 40, 41].

## 10. Random Forest Regressor (RFR)

Random Forest Regressor (RFR) is a "bagging" (bootstrap-aggregating) algorithm in which multiple trees are trained on random subsets of the training data so that the data is placed in different subsets. Two parameters of the known number of regression trees and the number of variables in each node specify the RFR method. This algorithm prioritizes the features regarding their importance, which weighs more on those data, mainly affecting the prediction model. To find the most influential variables and multivariate interaction, RFR uses the data permutation. In this technique, different trees are placed into different subsets, employing a dataset randomly for training. Random selection helps to avoid data deletion, while the specific data might be applied in many subsets or never be used in the training process [37, 38, 42].

## 11. Lasso

Least absolute shrinkage and selection operator, also known as lasso is a regression analysis technique used in statistics and machine learning that performs both variable selection and regularization in order to improve the accuracy of prediction and the interpretability of the results of the statistical model.

Lasso was initially developed for models of linear regression. This simple model exposes a lot about the estimator. These contain its relation to ridge regression and best subset selection and the association between lasso coefficient estimates and so-called soft thresholding. Additionally, it demonstrates that if covariates are collinear, the coefficient estimates do not necessarily need to be unique, unlike in standard linear regression. The details of this algorithm are explained in Ref. [43].

It should be mentioned that this paper used Lasso to regulate the classical machine learning algorithms, but deep learning methods extract features automatically.

## 12. Linear Regression

Linear regression analysis (LR) is statistical multivariate technique for studying of a relationship between a collection of independent variables (or predictors) and dependent variables (or criterion). Linear regression is a widely coupled chemometrics method with LIBS technique which pursues a linear predictor function as follow:

$$y = a_0 + a_1 X_1 + a_2 X_2 + \cdots + a_N X_N + \varepsilon \qquad (18)$$

here, $a_0$, $a_1$, $a_N$ are the regression coefficients, $\varepsilon$ is the residual error, and $y$ is the dependent variable so that this equation must be written for all samples.

In current paper, for designing a LR model, the spectral intensities at different concentrations are employed as independent variables, while elemental concentrations are considered as dependent parameters.

## 13. Data and features

In this paper, seven aluminum standards are selected for quantitative analysis. The certified concentrations of each aluminum alloy are represented in table 1. During this experiment, different spectral lines of Si (288.15 nm), Fe (274.93 nm), Cu (324.75 nm), Zn (334.40 nm), Mn (257.61 nm) and Mg (280.27, 285.21, and 279.55 nm) are observed which are chosen for statistical analysis. Here, three Mg spectral lines are appeared which the summation of them is used as the Mg intensity. After adjusting at optimum delay time, the experiments are performed.
For all samples, each spectrum is an average of 10 spectra so that 10 laser pulses strike a certain point of each sample. In addition, 87 average spectra are collected from whole Al alloys which mean that the proposed methodology used 870 spectra of different standard aluminum samples in correspondence of various concentrations of Si, Fe, Cu, Zn, Mn and Mg.
For prediction of the concentrations, 10 average spectral intensities related to various sample points containing different elements of the aluminum alloys are employed as inputs, and elements detected by LIBS method are used as features. Here, the details of concentrations, elements, and intensities for performing the predictions are summarized in table I in appendix.

Table 1: the characteristics of seven standard aluminum alloys with certified concentrations.

| Sample No | Si | Fe | Cu | Mn | Mg | Zn | Al |
|---|---|---|---|---|---|---|---|
| 1100 | 0.280 | 0.190 | 0.240 | 0.045 | 0.018 | 0.098 | 99.121 |
| 2017 | 0.895 | 0.582 | 3.430 | 0.225 | 0.650 | 0.029 | 94.147 |
| 3004 | 0.230 | 0.517 | 0.330 | 1.680 | 1.090 | 0.360 | 95.731 |
| 4043 | 4.550 | 0.750 | 0.300 | 0.070 | 0.250 | 0.180 | 93.737 |
| 5357 | 0.097 | 0.180 | 0.070 | 0.240 | 1.160 | 0.020 | 98.200 |
| 6010 | 0.930 | 0.410 | 0.360 | 0.440 | 1.030 | 0.320 | 96.211 |
| 7079 | 0.180 | 0.320 | 0.670 | 0.170 | 4.340 | 4.150 | 89.951 |

It should be noted that due to high concentrations of aluminum elements in all of standard samples, they are not used for prediction. Since, they are encountered to the self-absorption phenomenon. Therefore, the wavelengths of other elements with fewer concentrations are utilized. Here, all the chemometrics analysis including machine learning algorithms such as Recurrent Neural Networks and Convolutional Recurrent Networks, and as well as MLP, SVR, Decision Tree algorithm, Linear Regression, GBR, RFR, and k-Nearest Neighbor (KNN) algorithm were explained in detail in the above sections.

It can be noted that in all of calculations, python software is utilized and Tensorflow and Sicket-Learn toolbox is employed for the model developing. Here, one sample is selected for the validation stage, and the rest samples are used as training set. This has been repeated by randomly selecting one validation sample from the whole samples. Here, twenty percent of the data were used for testing, and eighty percent of the data were used for training, and the selections were made in a random state of 42.

It should be mentioned that in this research, to make the machine learning algorithms explainable, the Lasso method is employed. Figure 3 depicts the Lasso diagram of the effective value or coefficient of each input, including the intensities and the type of identified element on the prediction of aluminums concentration. As it is seen in this figure, Lasso results show that the third experiment has a great impact on aluminum samples and positive feedback in determining the aluminum concentration. The experiments of 6 and 8 also present that the taken intensities have a negative feedback and Lasso explains that the other experiments on the sample don't have a specific feedback for aluminum concentration.

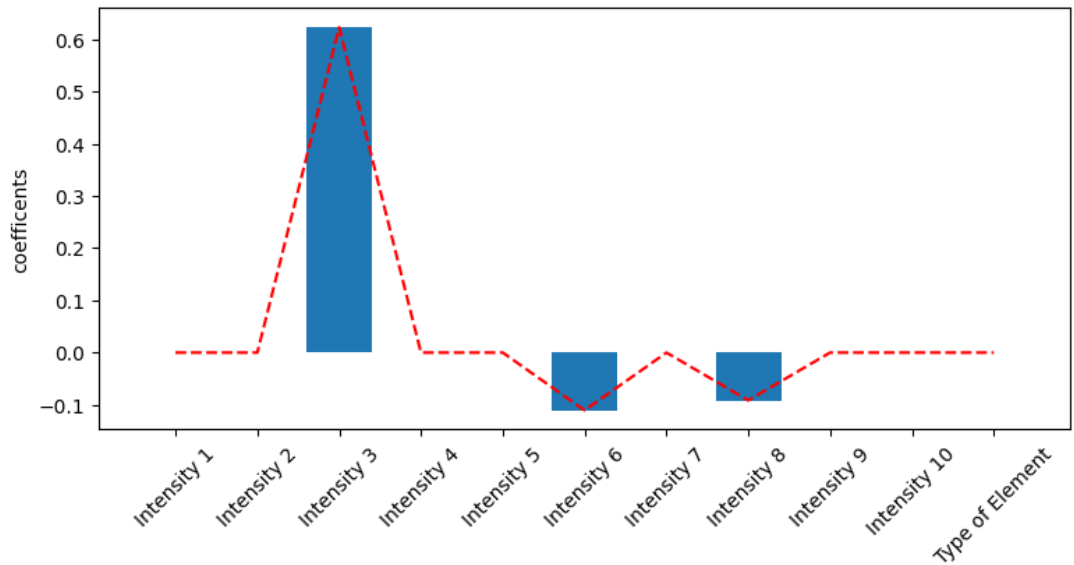

Figure 3. The evolution of the coefficient of each input versus the intensities and the type of identified element.

## 14. Errors estimation and accuracy evaluation

In statistics, the concept of error is an essential measurement used for evaluation of the performance of an estimator or predictor. Generally, there is different formula for calculation of errors. For instance, mean squared error (MSE) or mean squared deviation (MSD) assesses the quality of a predictor as follows [44, 45]:

$$MSE = \frac{1}{n}\sum_{i=1}^{n}(y_i - \hat{y}_i)^2. \quad (19)$$

In addition, for a better interpretation of the results, mean absolute error (MAE) can also be calculated by the average of all absolute errors as [46]:

$$MAE = \frac{1}{n}\sum_{i=1}^{n}|y_i - \hat{y}_i|. \tag{20}$$

The mean absolute percentage error (MAPE) is also used as one of the most commonly method to predict the accuracy [47]:

$$MAPE = \frac{1}{n}\sum_{i=1}^{n}|(y_i - \hat{y}_i)/y_i| \tag{21}$$

where, in Eqs. (19) to (21), $y_i$ and $\hat{y}_i$ are the target and estimated concentration magnitudes related to spectrum *i*, and *n* is the number of test spectra considered.

**Results and discussion**

The most frequent methods for composition prediction in LIBS technique are calibration curve and calibration free approaches. In this work, quantitative determinations are carried out by different deep learning methods like SimpleRNN, LSTM, GRU, and Conv-SimpleRNN, Conv-LSTM, Conv-GRU. Then, a comparison is performed with different classical multivariate methods of MLP, Linear Regression, Decision Tree algorithm, SVR, GBR, RFR, and KNN.

In Artificial Neural Network, the grid search method is employed for choosing the best transfer functions. Here, different transfer functions of purelin (linear), logsig (Log-sigmoid), and tansig (tangent sigmoid) [48,49] are tested for all of the samples in order to attain the best prediction with minimum errors [50]. The performance of the Recurrent Neural Networks model based on SimpleRNN, LSTM, and GRU are shown in figure 2. As it is clearly seen in this figure, the SimpleRNN predict the best results among other alghorithms, which is due to being in dynamical form which save the pervious states and help them for the next states.

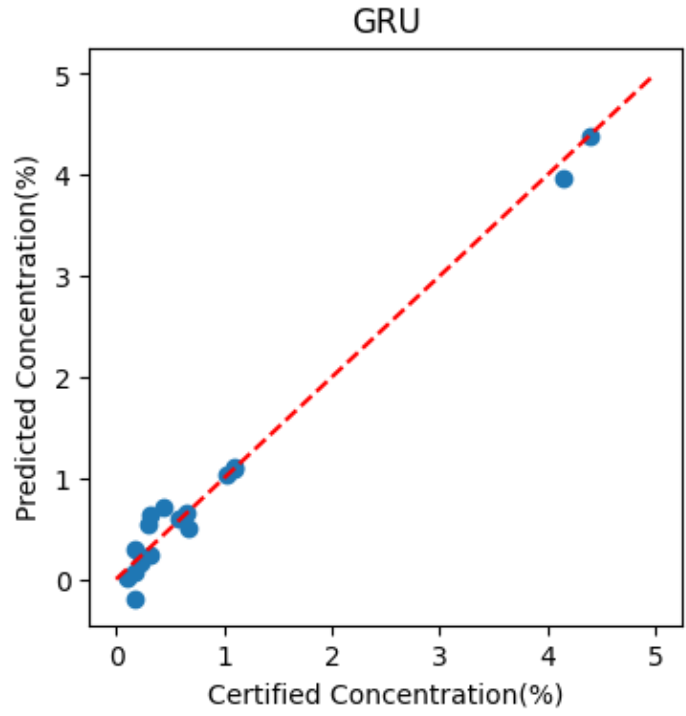

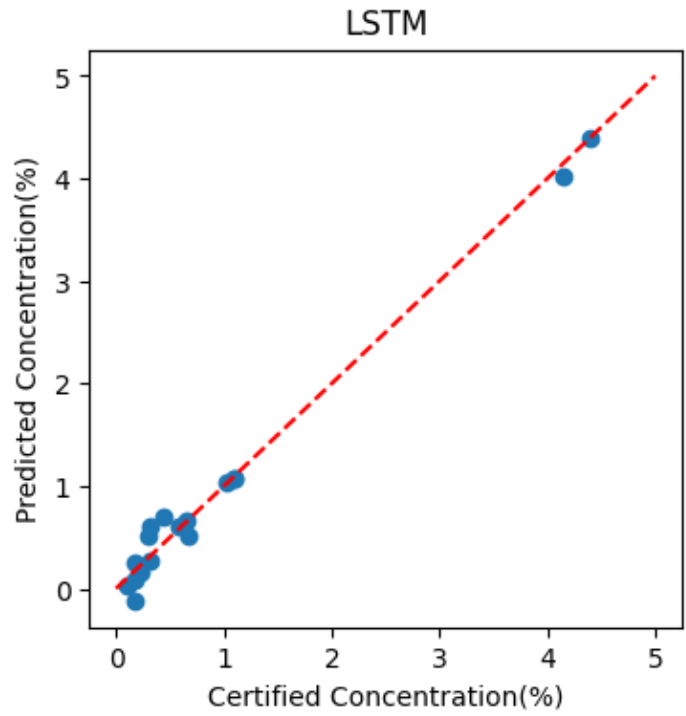

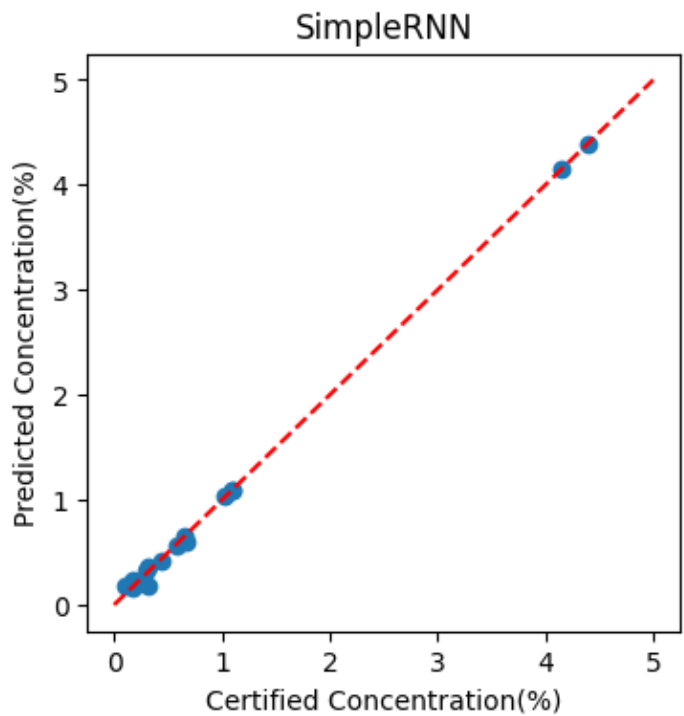

Figure 2. A comparison between concentration prediction of the whole elements by different Recurrent Neural Networks of SimpleRNN, LSTM, and GRU.

In final stage, Convolutional Recurrent Networks algorithms are added to different machine learning methods as Conv-SimpleRNN, Conv-LSTM, Conv-GRU, techniques. Figure 3 presents the concentration predictions by different Convolutional Recurrent Networks which shows considerable improvement in quantitative analysis. As it is clearly seen, the slope of correlation curve between the predicted and nominal concentration is near to 1 in all of represented deeplearning algorithms. As it is observed in these figures, Convolutional Recurrent Networks reported the nearest predicted values to certified concentrations. A feature which causes the application of the Convolutional Recurrent Networks methods in LIBS analysis induces much simpler (and more precise) with respect to other traditional algorithms is that Convolutional minimize preprocessing and they extract features from the data for the recurrent layer. Moreover, it should be mentioned that in all assimilation with Convolutional Recurrent Networks, Convolution layer with pooling layer is first applied to the input variables to extract useful features by data filtering.

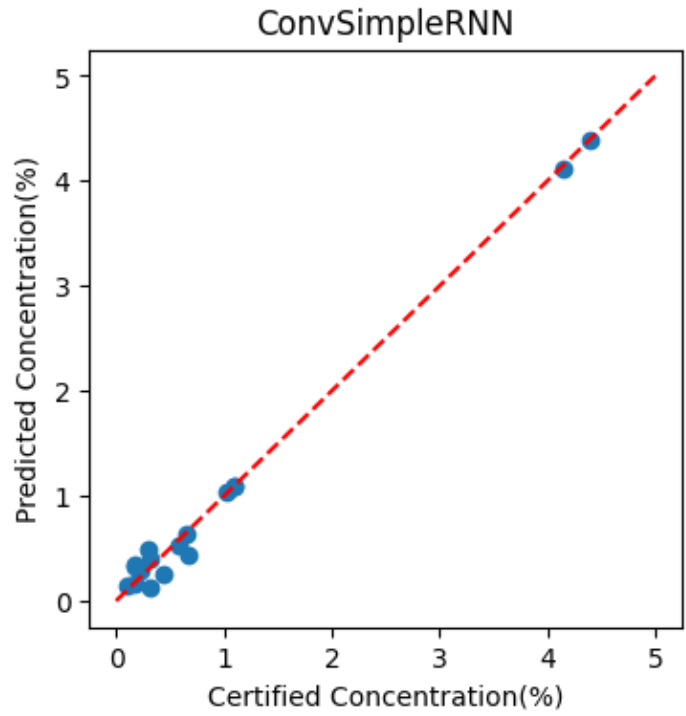

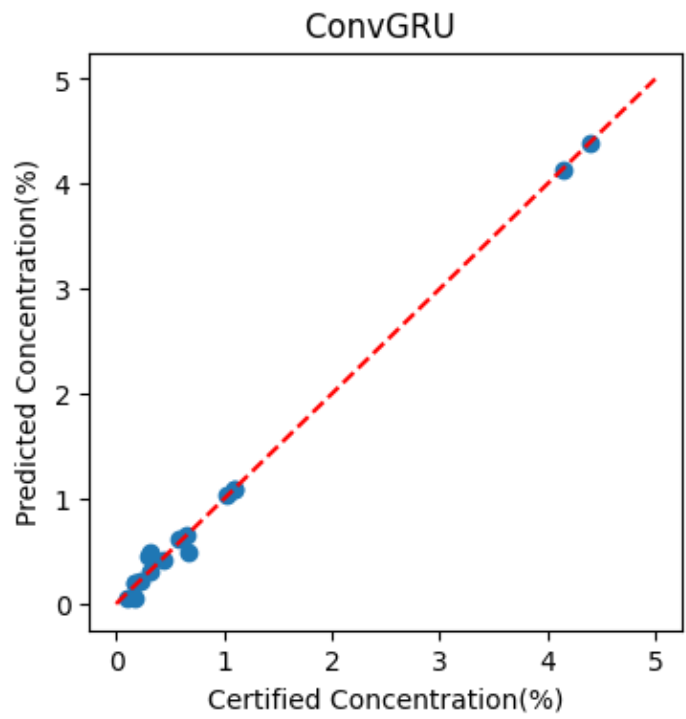

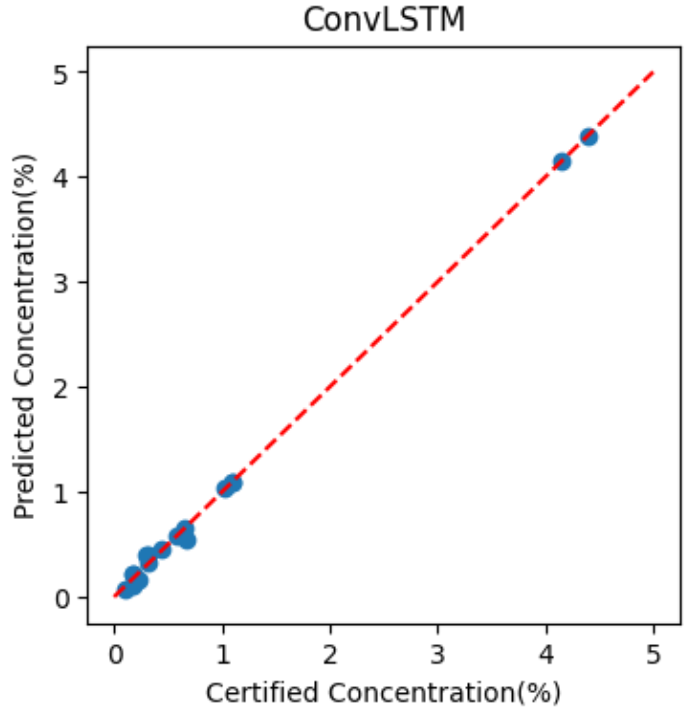

Figure 3. A comparison between concentrations prediction of the whole elements by Convolutional Recurrent Networks of Conv-SimpleRNN, Conv-LSTM, and Conv-GRU. As it is clearly seen, the slope of correlation curve between predicted and nominal concentration is near to 1 for all the elements.

As a final consideration, a comparison summarized in table 2 represents the prediction results of the concentration by different proposed approaches. The validity of these methods is estimated by comparing the statistical error values of mean MSE, mean MAPE and mean MAE calculated from constituent elements of standard aluminum samples.

Table 2. A comparison between different Recurrent Neural Networks in prediction of aluminum contents by calculation of mean squared error (MSE), the mean absolute error (MAE), and the mean absolute percentage error (MAPE).

| Networks | MSE | MAE | MAPE |
|---|---|---|---|
| **SimpleRNN** | 7.66e-5 | 0.00148 | 0.006739 |
| **Conv-SimpleRNN** | 7.668e-5 | 0.001486 | 0.006730 |
| **LSTM** | 0.0015580 | 0.00614 | 0.021179 |
| **ConvLSTM** | 0.0001444 | 0.001831 | 0.006406 |
| **GRU** | 0.0007582 | 0.004912 | 0.019347 |
| **ConvGRU** | 0.0003848 | 0.003233 | 0.011745 |

Furthermore, the results in table 3 prove the evidence that the proposed SimpleRNN and Conv-SimpleRNN as the best prediction methodologies have produced the significantly lowest MAE, MAPE, and RMSE (%) for all of elements. From table 3, it is found that the highest error value in SimpleRNN and Conv-SimpleRNN achieved to 0.006 % which reflect the robustness of this method. In addition, the comparison prediction results affirm that the Conv-SimpleRNN approach not only improves the prediction accuracy of the single LSTM or others methods, but also outperforms the competing methods in forecasting. As clearly seen in this table, in most of cases, integration with Recurrent Convolutional layer algorithms enhanced the accuracy of the single form of that special method except for SimpleRNN. Furthermore, taking into account the above considerations, the least mean squared error (MSE), mean absolute percentage error (MAPE), and the mean absolute error (MAE) of the prediction values are related to SimpleRNN method, and the worst one are seen in LSTM algorithm.

Generally, the integration of Convolutional Recurrent Networks and machine learning approaches illustrated good performance in different forecasting and classification fields, such as image manipulation detection [51], weather prediction [52], and text classification [53]. In this table, it is demonstrated that when Convolutional Recurrent Networks are not considered, SimpleRNNs approach can better predict the compositions with higher precisions rather than simple GRU, and LSTM. In conclusion, as it is obvious from table 3, the whole of Recurrent Neural Networks report good prediction in LIBS technique.

Here, the prediction models of linear regression, SVR, MLP, RFR, decision tree, k-nearest neighbor's, and gradient boosting are introduced as machine learning algorithms for making a comparison with Recurrent Neural Networks which use the composition and spectral intensities of aluminum's alloys for developing their performance.

A comparison of the analytical prediction of classical Machine learning algorithms is shown in figure 4 for different aluminum samples. In these figures it is seen that the feed-forward multilayer perceptron (MLP) with gradient descent algorithm shows the best prediction with relatively low values for MSE, MAE, and MAPE. Moreover, most of classical Machine learning methodologies show relatively high errors in forecasting the elemental's concentrations.

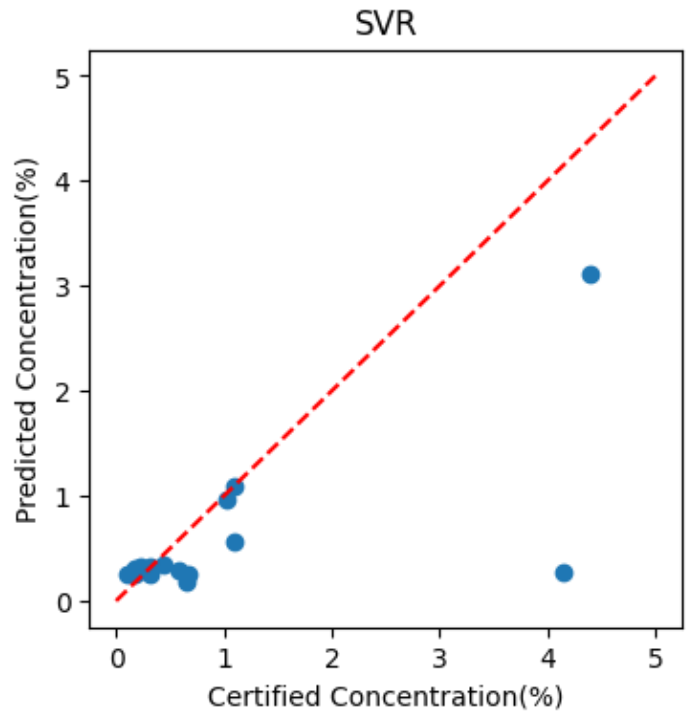

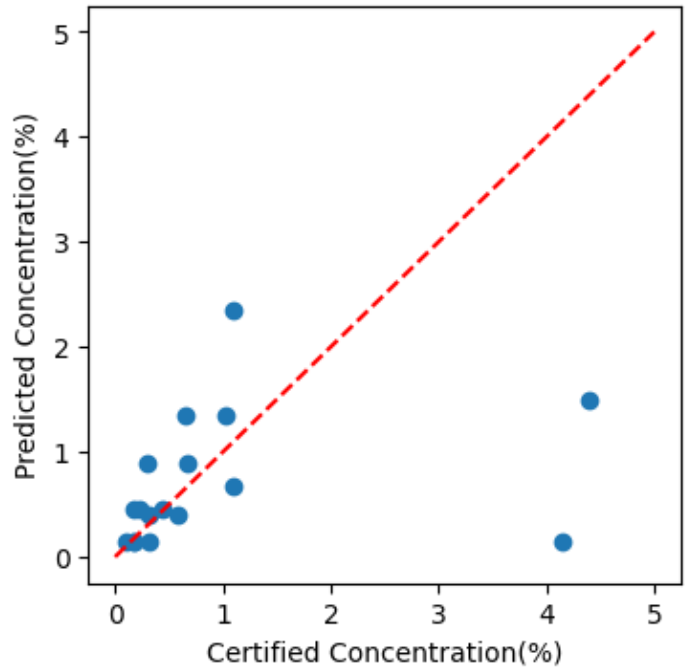

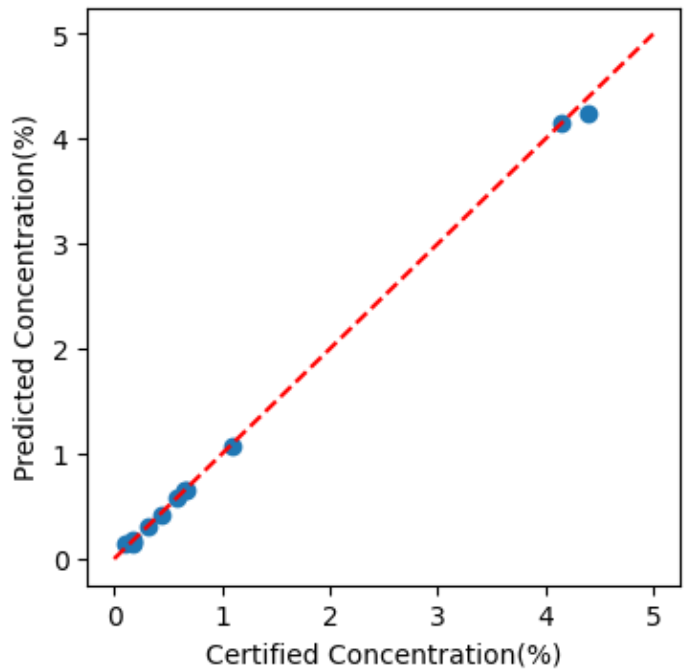

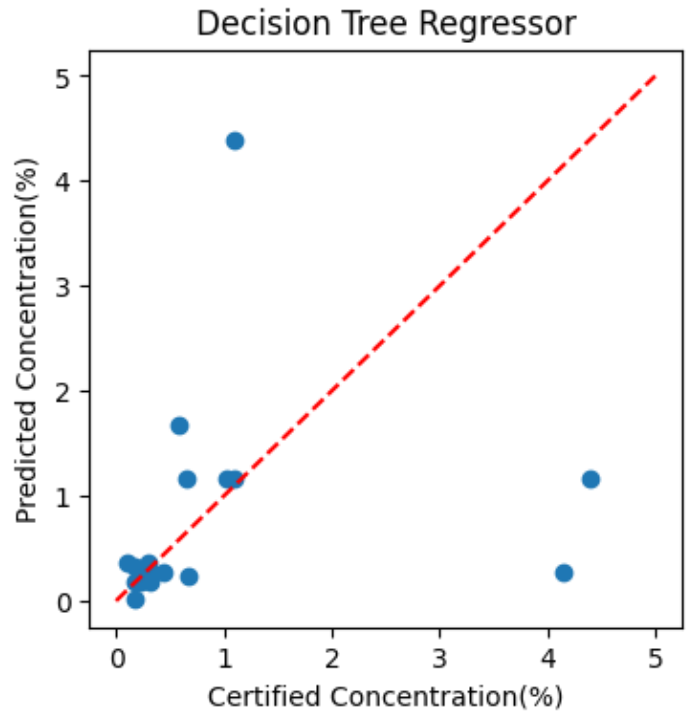

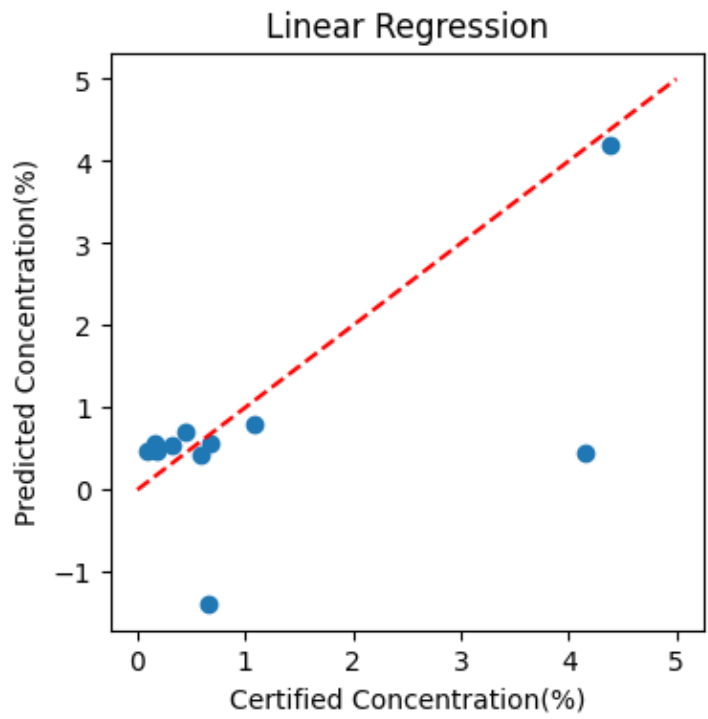

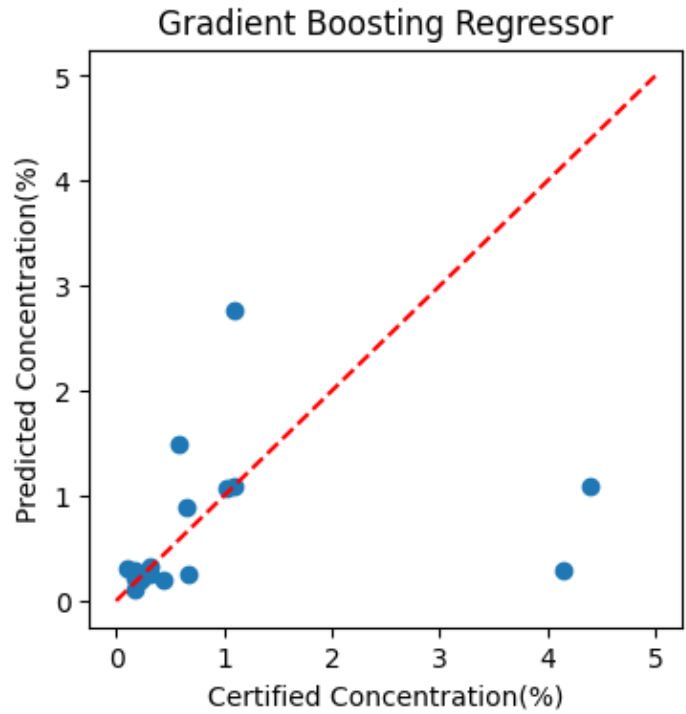

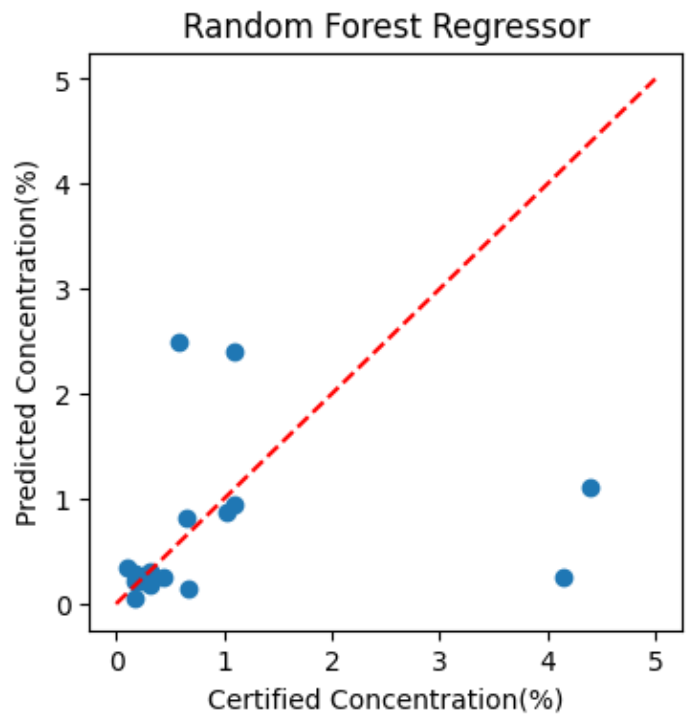

Figure 4. A comparison among predictions by different classical machine learning algorithms.

It can be noted that another approach to predict the composition of the standard aluminum alloys is using kernel-based support vector regression (KSVR). In this research, the influence of different kernel functions on composition prediction is examined and their performance is compared in terms of MSE, MAPE, and MAE. It is seen that the linear kernel or simple SVR almost forecasts better results rather than other kernels functions in different elements, but in current research it is not as good as MLP algorithm in prediction process.

By making a comparison between the reports of the classical machine learning algorithm with prediction by RNNs, it is seen that the accuracies of these methods are not very high and these are not recommended for quantitative purpose except MLP.

After examination of different classical methods, it is clearly seen that in the most of approaches, sometimes great deviations are seen at high concentrations which is due to self-absorption effects which causes the multivariate methods could not perform well in prediction of aluminum's composition [54]. It should be stressed that in our previous paper [23], the self-absorption effect was corrected which caused the Neural Networks predictions improved considerably.

Results of modeling with Machine learning algorithms are summarized in table 3 by illustration of the related error bars of MSE, MAPE, and MAE for showing the precision of measurements. Red colors in this table show the lowest values of errors predicted by different algorithms. Consequently, it can be concluded that the MLP method is able to reproduce the concentrations with acceptance standard errors.

Table 3. A comparison between various classical Machine learning algorithms in prediction of aluminum concentrations by calculation of mean squared error (MSE), the mean absolute error (MAE), and the mean absolute percentage error (MAPE).

| Algorithm | MSE | MAE | MAPE |
|---|---|---|---|
| Linear Regression | 0.12736 | 0.0617 | 0.1312 |
| SVR | 0.12473 | 0.0512 | 0.0455 |
| MLP | 0.00053 | 0.0029 | 0.0035 |
| Random Forest | 0.07839 | 0.0355 | 0.0338 |
| Decision Tree | 0.16379 | 0.0540 | 0.0577 |
| k-nearest neighbors | 0.09008 | 0.0379 | 0.0428 |
| Gradient Boosting | 0.09759 | 0.0386 | 0.0397 |

**Conclusion**

The determination of the exact composition of the aluminum alloys is somewhat difficult by LIBS, due to the important matrix effects and the nonlinearity relation of the spectral intensities versus concentration. In this paper, different recurrent deep learning methods such as SimpleRNN, GUR, LSTM, and Convolutional Recurrent Networks (Conv-SimpleRNN, Conv-LSTM, Conv-GRU) are employed for quantitative analysis. Furthermore, support vector regressor (SVR), Linear Regression, the multilayer perceptron (MLP), decision tree algorithm, gradient boosting regression (GBR), random forest regression (RFR), and k-nearest neighbor (KNN) algorithm as classical Machine learning methodological approaches are coupled with LIBS technique in order to evaluate the effectiveness of the proposed models and introduce the best quantitative methods in comparison with Recurrent algorithms. For each mentioned algorithm, the general variation of the errors including MSE, MAPE, and MAE is reported. It was seen that in most of samples, the Convolutional Recurrent Networks in the SimpleRNNs improved significantly the accuracy compared to the other algorithms. The experimental results verified that Convolutional Recurrent Networks is preferable methods with respect to the other deep learning approaches only through the comparison of different error values. Although classical algorithms do not have very high accuracy in modeling like neural networks, but with the help of regulation methods of classical algorithms such as Lasso, the explainability of modeling can be understood, but in neural networks, this clear explainability is not existed compared to modeling methods based on experiments. Finally, a definite benefit of the proposed approaches is the possibility of using them effectively for giving information on constituent elements of each arbitrary sample in LIBS analysis.

**Acknowledgement**

The authors are thankful from Professor Seyed Hassan Tavassoli for sharing his laboratories equipment and his kind supports.

**Appendix**

In this research, the details of concentrations, elements, and intensities for accomplishing the predictions are summarized in table I. In these tables, Intensity 1 to Intensity 10 means that the average magnitudes of different spectral intensities related to 10 sample points containing various elements at different wavelengths are presented.

Table I: Various concentrations, elements, and intensities for constructions of Neural Networks.

| Certified Concentration (%) | Intensity 1 | intensity 2 | intensity 3 | intensity 4 | intensity 5 | intensity 6 | intensity 7 | intensity 8 | intensity 9 | intensity 10 | Element |
|---|---|---|---|---|---|---|---|---|---|---|---|
| 0.098 | 316.190 | 262.019 | 473.325 | 482.161 | 408.012 | 515.970 | 434.137 | 312.732 | 373.819 | 315.422 | Zn |
| 0.029 | 218.221 | 274.313 | 302.743 | 233.973 | 306.585 | 288.528 | 354.993 | 263.940 | 250.109 | 646.980 | Zn |
| 0.360 | 383.039 | 415.312 | 474.093 | 373.819 | 391.107 | 536.332 | 427.222 | 379.582 | 404.170 | 438.747 | Zn |
| 0.180 | 353.072 | 411.470 | 333.863 | 577.057 | 389.187 | 476.782 | 431.448 | 400.712 | 297.749 | 327.332 | Zn |
| 0.020 | 465.641 | 371.898 | 350.767 | 426.069 | 335.015 | 424.532 | 372.666 | 358.835 | 237.815 | 308.122 | Zn |
| 4.150 | 374.971 | 1152.962 | 1089.570 | 1367.725 | 1458.010 | 1359.273 | 1147.967 | 1323.159 | 1699.667 | 1366.572 | Zn |
| 0.320 | 563.994 | 473.709 | 441.821 | 616.628 | 474.093 | 487.156 | 508.670 | 1378.098 | 466.793 | 426.453 | Zn |
| 0.045 | 3061.234 | 1597.296 | 2295.551 | 2742.574 | 1206.213 | 1221.697 | 2233.118 | 3446.823 | 2762.054 | 1996.370 | Mn |
| 0.225 | 1411.994 | 1360.549 | 1202.218 | 1191.229 | 1412.493 | 662.293 | 1224.194 | 591.369 | 1380.028 | 1703.183 | Mn |
| 1.680 | 1708.178 | 1719.666 | 1675.213 | 2460.375 | 1159.763 | 2142.215 | 1304.608 | 1809.070 | 2480.854 | 2457.878 | Mn |
| 0.070 | 1680.208 | 1334.576 | 928.509 | 1730.154 | 1648.242 | 1553.842 | 866.076 | 1799.081 | 4012.720 | 2971.331 | Mn |
| 0.240 | 1458.944 | 1170.252 | 1082.845 | 1626.265 | 1876.998 | 1322.090 | 1647.243 | 863.578 | 732.718 | 3446.823 | Mn |
| 0.170 | 1302.111 | 606.852 | 1273.142 | 712.240 | 1113.812 | 785.162 | 898.042 | 707.245 | 976.458 | 1751.132 | Mn |
| 0.440 | 1924.947 | 886.554 | 1083.844 | 2101.758 | 1608.784 | 1946.424 | 2090.271 | 1124.301 | 1202.218 | 2080.281 | Mn |
| 0.240 | 1014.158 | 897.481 | 669.780 | 854.686 | 951.177 | 989.127 | 882.947 | 928.972 | 947.139 | 1000.835 | Cu |
| 3.430 | 1525.275 | 1918.100 | 1444.933 | 1449.778 | 1754.591 | 2034.373 | 2660.147 | 2025.895 | 1508.722 | 2425.986 | Cu |
| 0.330 | 336.707 | 771.519 | 258.384 | 513.942 | 455.806 | 642.731 | 304.812 | 491.333 | 448.942 | 467.917 | Cu |
| 0.300 | 724.687 | 337.918 | 560.774 | 1202.698 | 582.172 | 631.022 | 279.378 | 268.477 | 1201.083 | 1017.792 | Cu |
| 0.070 | 186.924 | 173.198 | 267.670 | 275.340 | 254.347 | 263.632 | 335.899 | 274.937 | 163.912 | 869.220 | Cu |
| 0.670 | 474.377 | 659.687 | 864.780 | 637.078 | 857.109 | 442.483 | 457.824 | 549.874 | 2585.05 | 714.997 | Cu |
| 0.360 | 449.346 | 430.775 | 536.147 | 552.296 | 669.377 | 792.513 | 855.090 | 349.626 | 414.222 | 517.172 | Cu |
| 0.280 | 3479.575 | 2303.816 | 1123.199 | 2729.156 | 1480.078 | 422.690 | 878.948 | 1023.378 | 907.215 | 933.716 | Si |
| 0.895 | 1952.678 | 2492.415 | 3092.219 | 2700.005 | 2480.931 | 1336.531 | 2426.162 | 761.460 | 1051.646 | 1475.661 | Si |
| 0.230 | 323.753 | 729.217 | 610.846 | 1378.491 | 973.026 | 951.384 | 1451.811 | 499.984 | 917.374 | 652.365 | Si |
| 4.550 | 6953.850 | 4964.070 | 3889.015 | 6835.038 | 5008.238 | 7526.712 | 5709.630 | 244.692 | 12772.580 | 11498.320 | Si |
| 0.097 | 422.248 | 455.816 | 459.791 | 555.636 | 512.793 | 432.407 | 624.980 | 274.726 | 292.393 | 765.435 | Si |
| 0.180 | 1245.103 | 1071.080 | 878.064 | 729.217 | 442.565 | 447.424 | 489.825 | 464.649 | 1486.262 | 1067.547 | Si |
| 0.930 | 2259.648 | 1028.679 | 1385.116 | 3829.829 | 1838.283 | 3394.331 | 6363.762 | 903.682 | 781.778 | 1460.644 | Si |
| 0.018 | 16851.146 | 10991.425 | 11853.185 | 17809.466 | 11444.508 | 232125.87 | 10716.269 | 10041.967 | 10292.861 | 10276.444 | Mg |
| 0.650 | 31644.717 | 34098.631 | 22182.74 | 33117.336 | 47032.388 | 67816.18 | 126219.86 | 78561.805 | 80118.748 | 71668.301 | Mg |
| 1.090 | 62023.574 | 64539.35 | 75499.074 | 149048.682 | 59871.804 | 144481.024 | 120623.024 | 131659.362 | 169302.99 | 197056.92 | Mg |
| 0.250 | 24293.757 | 22938.243 | 30672.428 | 36515.511 | 27624.576 | 47500.270 | 33667.860 | 185665.69 | 41917.400 | 46365.905 | Mg |
| 1.160 | 46549.483 | 67713.182 | 97734.763 | 71724.571 | 122933.176 | 126012.642 | 165566.778 | 128445.955 | 101874.803 | 144687.400 | Mg |
| 4.390 | 100894.648 | 62335.73 | 159684.84 | 134512.8 | 96640.512 | 54371.483 | 82788.15 | 92948.03 | 127553.944 | 198249.22 | Mg |
| 1.030 | 84699.079 | 80979.841 | 54954.889 | 51390.995 | 45114.971 | 35457.896 | 37590.724 | 134577.581 | 129538.161 | 120826.330 | Mg |
| 0.190 | 1530.424 | 844.079 | 3736.825 | 3781.063 | 2354.246 | 909.361 | 631.356 | 762.788 | 644.718 | 635.271 | Fe |
| 0.582 | 975.460 | 609.071 | 597.905 | 842.038 | 2189.064 | 2637.533 | 1489.704 | 829.850 | 2410.728 | 2022.350 | Fe |
| 0.517 | 2495.121 | 2828.675 | 3114.237 | 2921.697 | 2188.524 | 2510.694 | 1762.119 | 1554.622 | 1299.690 | 1505.292 | Fe |
| 0.750 | 2669.283 | 3125.012 | 4960.184 | 4956.810 | 4366.445 | 3603.951 | 2372.376 | 3426.975 | 3089.659 | 2602.944 | Fe |
| 0.180 | 2737.363 | 2172.434 | 1261.031 | 2523.076 | 1134.198 | 1939.567 | 1835.604 | 831.562 | 773.407 | 989.706 | Fe |
| 0.320 | 1827.014 | 1989.893 | 1543.233 | 1426.326 | 1624.295 | 1308.388 | 1236.243 | 1386.408 | 1361.749 | 1588.361 | Fe |
| 0.410 | 3283.010 | 2193.647 | 2505.571 | 3432.501 | 4049.765 | 3651.207 | 2541.616 | 2008.822 | 1751.850 | 1358.266 | Fe |

## Author contributions

Pouriya Khaliliyan contributed to data analysis by performing the statistical calculations, and Fatemeh Rezaei supervised and prepared initial version of paper by writing the manuscript. Parvin Karimi performed the experiments. Mohsen Rezaei and Behnam Ashrafkhani contributed on machine learning sections of manuscript.

## Competing interests

The author(s) declared no potential conflicts of interest with respect to the research, authorship, and/or publication of this article.

**Data availability statement**

The data that support the findings of this study are available from the corresponding author upon reasonable request.